\documentclass[11pt,a4paper]{article}

\PassOptionsToPackage{hyphens}{url}
\usepackage[utf8]{inputenc}
\usepackage[T1]{fontenc}
\usepackage{amsmath,amssymb}
\usepackage{booktabs}
\usepackage{graphicx}
\usepackage{hyperref}
\usepackage[margin=2.5cm]{geometry}
\usepackage{enumitem}
\usepackage{xcolor}
\usepackage{natbib}
\usepackage{listings}
\usepackage{tabularx}
\usepackage{multirow}
\usepackage{float}
\usepackage{placeins}

\graphicspath{{./}}

\hypersetup{
    colorlinks=true,
    linkcolor=blue!70!black,
    citecolor=blue!70!black,
    urlcolor=blue,
    breaklinks=true
}
\makeatletter
\g@addto@macro{\UrlBreaks}{\UrlOrds}
\makeatother

\newcommand{\model}{\texttt{gpt-oss-120b}}
\newcommand{\nemotron}{\texttt{Nemotron-Super-120B}}
\newcommand{\qwen}{\texttt{Qwen3.5-35B-A3B}}
\newcommand{\gpu}{H100 80\,GB}

\title{Model Capability Dominates: Inference-Time Optimization Lessons from AIMO~3}

\author{
    Natapong Nitarach \\
    \texttt{natapong.nitarach@proton.me}
}

\date{Apr 2026}

\begin{document}

\maketitle

\begin{abstract}
Majority voting over multiple LLM attempts improves mathematical reasoning, but correlated errors limit the effective sample size. A natural fix is to assign different reasoning strategies to different voters. The approach, \textit{Diverse Prompt Mixer}, is tested on the AIMO~3 competition \citep{ai-mathematical-olympiad-progress-prize-3}: 3 models, 23+ experiments, 50 IMO-level problems, one \gpu{}, 5-hour limit. Every prompt-level intervention fails. High-temperature sampling already decorrelates errors; weaker strategies reduce accuracy more than they reduce correlation. Across an 8-point capability gap at equal $N{=}8$ and every optimization tested, model capability dominates. The gap between the best majority-vote score (42/50) and $\text{pass@}20$ (${\approx}45.5$) is selection loss, not prompt loss. A verifier-based selector could close it. Prompt engineering cannot. Code and notebooks: \url{https://github.com/nat-nischw/model-capability-dominates-lessons-aimo3}
\end{abstract}

\section{Introduction}

Majority voting across $N$ inference attempts (self-consistency; \citealp{wang2023selfconsistency}) is the standard approach for competition mathematics. It rests on the Condorcet Jury Theorem \citep{condorcet1785}: with per-attempt accuracy $p > 0.5$ and independent errors, the majority converges to the correct answer as $N$ grows.

In practice, errors are correlated. The same model with the same prompt makes the same systematic mistakes regardless of random seed. The effective sample size becomes
\begin{equation}
    N_{\text{eff}} = \frac{N}{1 + (N-1)\rho}
    \label{eq:majority}
\end{equation}
where $\rho$ is the mean pairwise error correlation. At $\rho = 0.3$, eight attempts reduce to 2.6 effective votes.

The natural fix: assign different reasoning strategies to different voters. If ``small cases first'' and ``work backwards'' lead the model through different paths, their errors should be less correlated. Call this \textit{Diverse Prompt Mixer}.

\begin{figure}[t]
\centering
\includegraphics[width=0.85\linewidth]{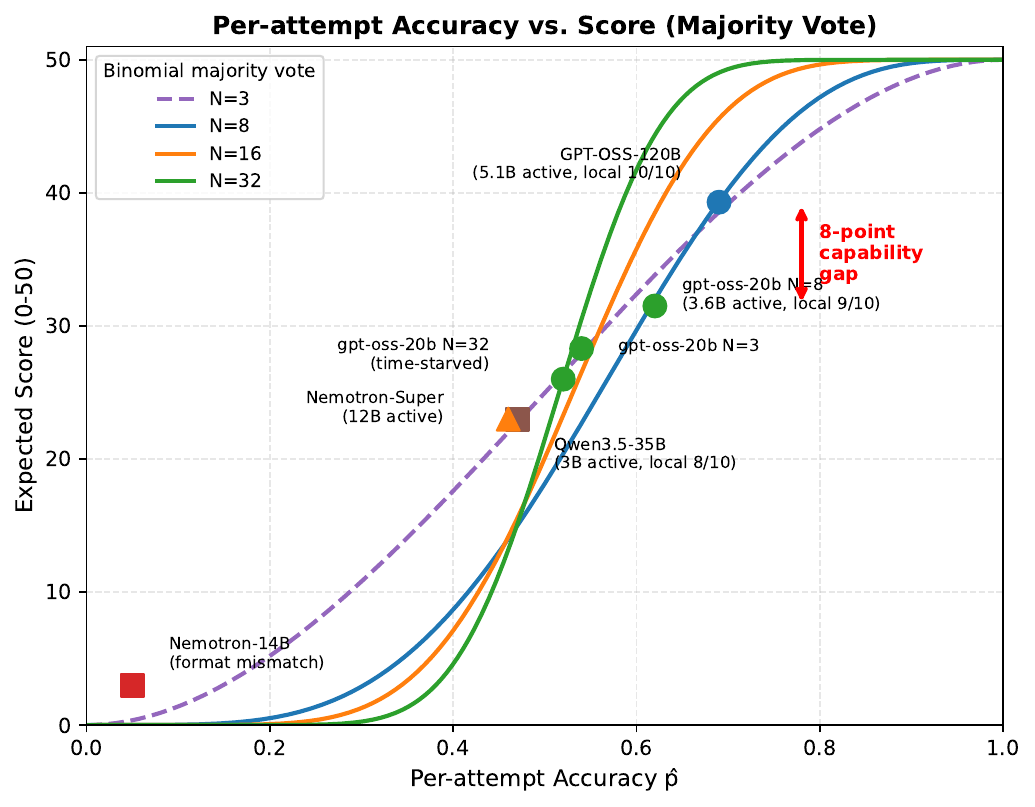}
\caption{\textbf{Model capability dominates.} Per-attempt accuracy $\hat{p}$ vs.\ expected majority-vote score under Binomial voting at $N{=}3,8,16,32$. Seven data points across four model families. At equal $N{=}8$, the 8-point gap between \model{} ($\hat{p}{=}0.69$, score 39.3) and \texttt{gpt-oss-20b} ($\hat{p}{=}0.61$, score 31.0) dwarfs every prompt optimization ($\pm$2 points). Scaling $N$ beyond compute budget backfires: \texttt{gpt-oss-20b} drops from 31.0 ($N{=}8$) to 26 ($N{=}32$) as per-attempt time shrinks.}
\label{fig:p-vs-score}
\end{figure}

The idea is intuitive. It is also wrong. This paper explains why (Figure~\ref{fig:p-vs-score}).

\section{System Architecture}
\label{sec:system}

The system runs on a single \gpu{} within Kaggle's 5-hour wall-clock limit. No external APIs, no multi-GPU setups, no pre-computed solutions.

\subsection{Model and Serving}

The model is \model{} \citep{gptoss2025modelcard}: 116.8B total parameters, 5.1B active via Mixture-of-Experts (MoE), served locally via vLLM \citep{kwon2023vllm} with FP8 quantization for both weights and KV cache. Weights consume ${\sim}75$\,GB; the remaining ${\sim}5$\,GB supports KV cache for $N{=}8$ concurrent sequences at 65536-token context.

Server startup takes ${\sim}160$\,s. Weights are pre-loaded via sequential file reads before launching vLLM to avoid cold-start latency.

\subsection{Inference Pipeline}

Each problem passes through a 5-stage pipeline:

\begin{enumerate}[nosep]
    \item \textbf{Budget allocation}: Divide remaining wall-clock time equally among remaining problems. No bonus, no inflation. This guarantees total solving time $\leq$ notebook limit.
    \item \textbf{Parallel attempts}: Launch $N{=}8$ concurrent inference threads, each with a different random seed but identical system prompt and temperature $T{=}1.0$.
    \item \textbf{Tool-integrated reasoning}: Each attempt can generate Python code via \texttt{<tool\_call>} tags. Code executes in a persistent Jupyter kernel sandbox with access to \texttt{sympy}, \texttt{numpy}, \texttt{mpmath}. Results feed back into the conversation for multi-turn reasoning.
    \item \textbf{Early stopping}: If 4 of 8 attempts agree on a non-trivial answer ($>1$), remaining attempts are cancelled.
    \item \textbf{Entropy-weighted voting}: Final answer selected by weighted majority vote, where weight $w = 1 + 1/(\text{entropy} + 0.1)$. Low-entropy (confident) attempts contribute more.
\end{enumerate}

\subsection{Time Management}

The 5-hour limit is the binding constraint. The budget system:

\begin{itemize}[nosep]
    \item Total: 18000\,s. Reserve 540\,s for infrastructure variance, 360\,s for startup.
    \item Solving budget: 17100\,s for 50 problems.
    \item Per-problem budget: $\text{time\_left} / \text{problems\_remaining}$. Pure equal division; no bonus from saved time. Mathematically guarantees completion.
    \item Hard deadline: if $<$30\,s remain, return 0 immediately. This ensures the inference server answers all 50 problems before Kaggle's hard kill.
\end{itemize}

\section{Diverse Prompt Mixer}

Four complementary strategies, each a different system prompt with all other parameters identical:

\begin{enumerate}[nosep]
    \item \textbf{Original}: Step-by-step with exploration, planning, and verification.
    \item \textbf{Small Cases First}: Enumerate $n{=}1,2,3,\ldots$, discover patterns, conjecture, prove.
    \item \textbf{Work Backwards}: List constraints, narrow search space, construct.
    \item \textbf{Classify Then Solve}: Identify problem type, recall canonical techniques, apply.
\end{enumerate}

Three ensemble configurations plus isolated strategies:

\begin{table}[htbp]
\centering
\begin{tabular}{lcccccc}
\toprule
Configuration & Orig & Small & Back & Class & $N$ & LB Score \\
\midrule
Baseline (21 runs) & 8 & 0 & 0 & 0 & 8 & 39.3 mean \\
Conservative & 5 & 1 & 1 & 1 & 8 & 40 \\
Aggressive & 3 & 2 & 2 & 1 & 8 & 40 \\
Equal & 2 & 2 & 2 & 2 & 8 & 38 \\
\midrule
\multicolumn{7}{l}{\textit{Isolated strategies (8$\times$ each):}} \\
8$\times$ Small Cases & 0 & 8 & 0 & 0 & 8 & 37 \\
8$\times$ Work Backwards & 0 & 0 & 8 & 0 & 8 & 39 \\
8$\times$ Classify & 0 & 0 & 0 & 8 & 8 & 36 \\
8$\times$ Code-first & \multicolumn{4}{c}{(code-first strategy)} & 8 & 41, 38, 34 \\
Formalize-First (F-1) & \multicolumn{4}{c}{(equations before code)} & 8 & 39 \\
\bottomrule
\end{tabular}
\caption{Prompt diversity configurations on \model{}. More diversity = worse performance. Code-first: 41, 38, 34 across three runs (mean 37.7, below baseline).}
\label{tab:diversity}
\end{table}

\begin{figure}[t]
\centering
\includegraphics[width=0.85\linewidth]{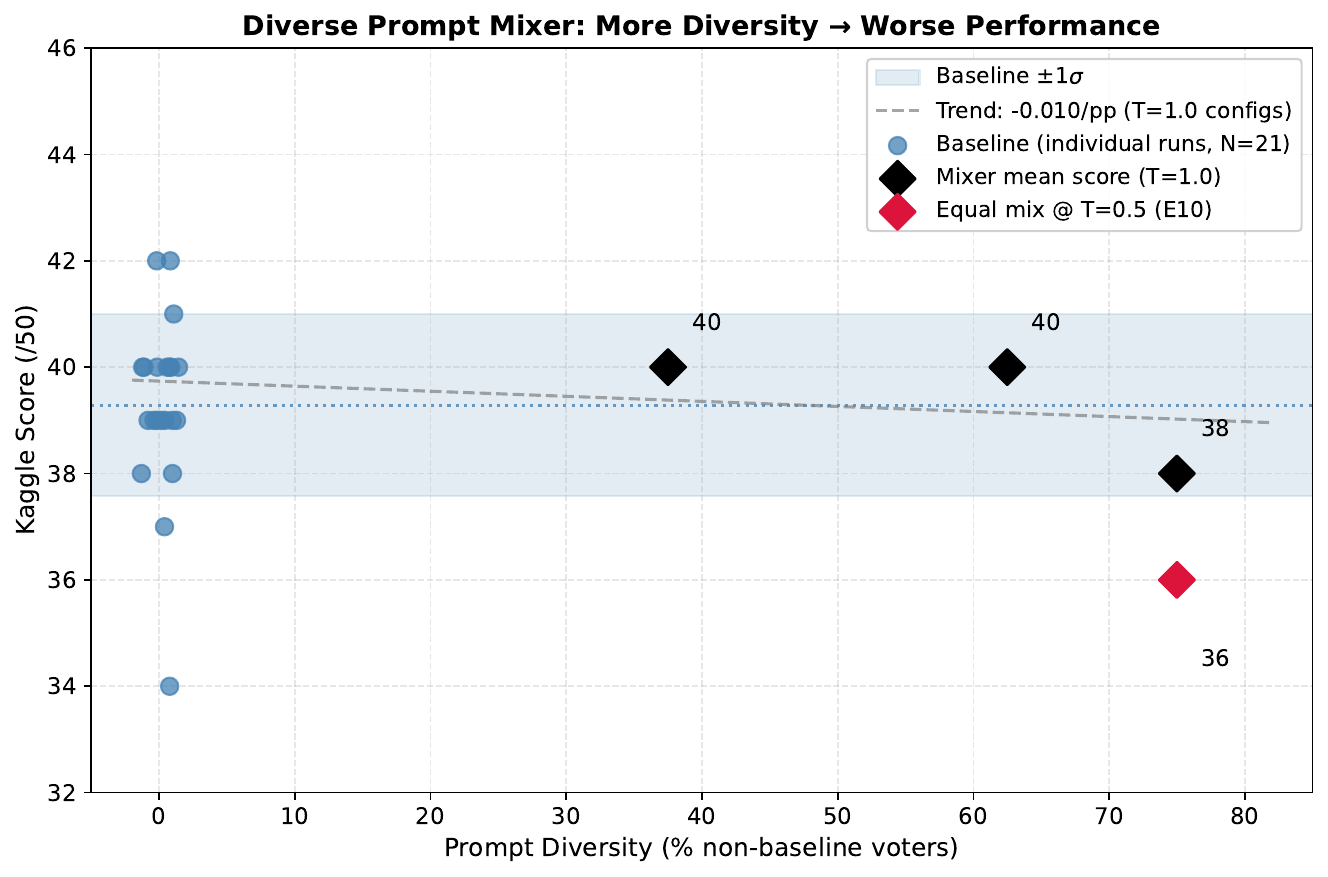}
\caption{Prompt diversity vs.\ score on \model{}. Blue circles: individual baseline runs ($N{=}21$). Black diamonds: configuration means. Shaded band: baseline $\pm 1\sigma$. More diversity monotonically degrades performance.}
\label{fig:diversity}
\end{figure}

More diversity = worse performance (Figure~\ref{fig:diversity}). The relationship is monotonic: replacing Original prompts with diverse strategies never helps and eventually hurts. Every individual strategy is weaker than the Original. Code-first (41/50) appeared promising but confirmation runs scored 38 and 34 (3-run mean 37.7, below baseline).

\section{Why It Fails}

\subsection{Temperature Already Decorrelates}

At $T{=}1.0$, the model explores different reasoning paths even with identical prompts. Temperature ablation confirms $T{=}1.0$ is optimal:

\begin{table}[htbp]
\centering
\begin{tabular}{lcl}
\toprule
Temperature & LB Score & $\Delta$ from baseline \\
\midrule
$T{=}0.5$ & 38 & $-1.3$ \\
$T{=}0.8$ & 40 & $+0.7$ \\
$T{=}1.0$ & 39.3 & --- (baseline, 21-run mean) \\
$T{=}1.2$, min\_p$=$0.03 & 37 & $-2.3$ \\
\bottomrule
\end{tabular}
\caption{Temperature ablation on \model{}. $T{=}1.0$ is optimal; both lower and higher temperatures degrade performance.}
\label{tab:temperature}
\end{table}

Prompt diversity on top of high temperature is redundant. Stochastic diversity already achieves most of the achievable decorrelation.

\subsection{Pairwise Correlation Is Already Near Zero}

Equation~\ref{eq:majority} assumes constant pairwise correlation $\rho$ across attempts. The method-of-moments estimator for a binary exchangeable sequence gives $\hat{\rho}$ directly:
\[
\hat{\rho} = \frac{v_c(v_c-1)/[N(N-1)] - \hat{p}^2}{\hat{p}(1-\hat{p})}
\]
where $v_c$ is the number of correct votes and $\hat{p} = v_c/N$. Computed across all four models: \model{} ($N{=}8$), \texttt{gpt-oss-20b} ($N{=}3$ and $N{=}8$), Qwen3.5-35B-A3B ($N{=}16$), and Nemotron-Super-120B ($N{=}3$). Problems with $\hat{p}\in\{0,1\}$ are excluded (undefined $\hat{\rho}$).

\begin{figure}[htbp]
\centering
\includegraphics[width=0.82\linewidth]{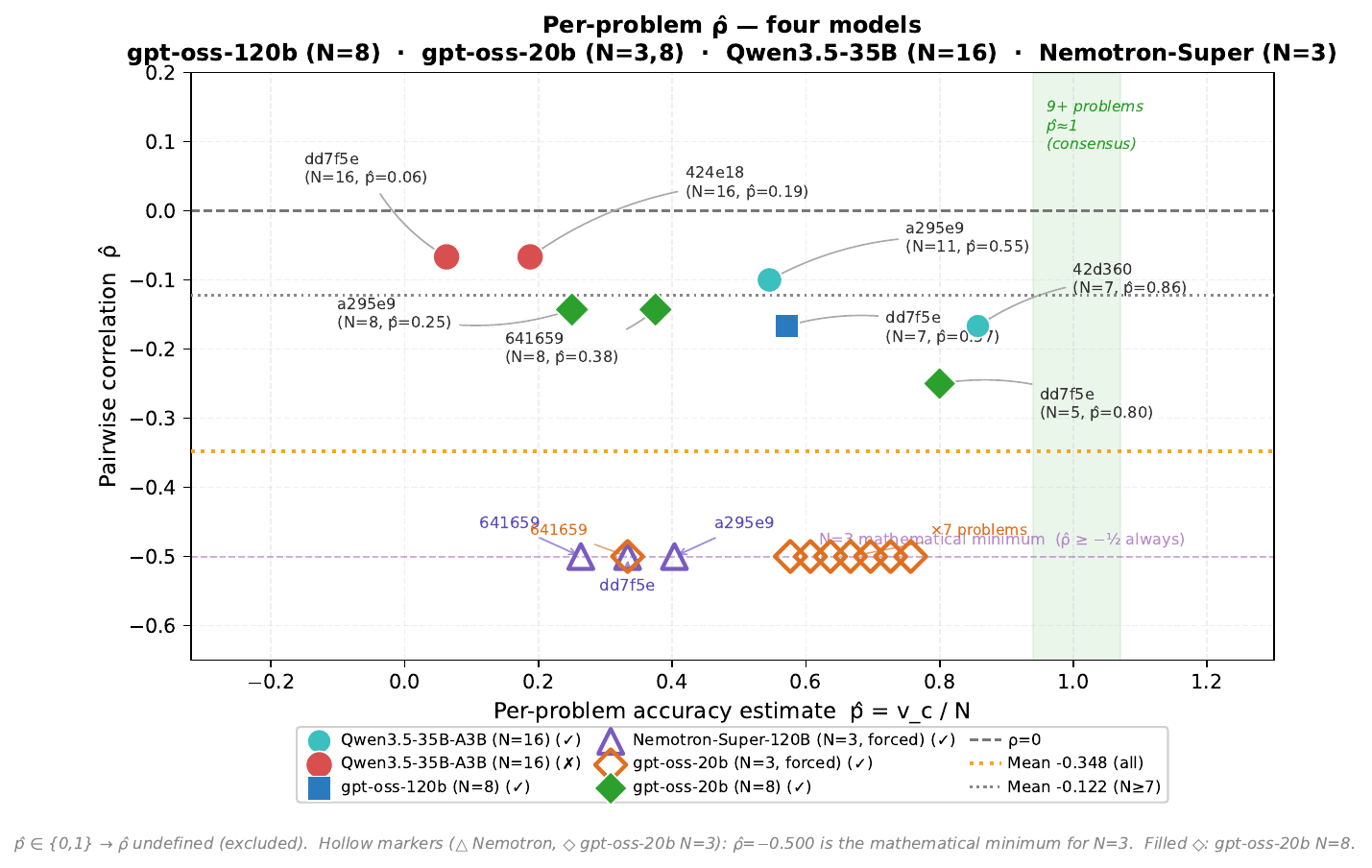}
\caption{Per-problem $\hat{\rho}$ vs.\ $\hat{p}$ across four models. Circles: Qwen ($N{=}16$); squares: \model{} ($N{=}8$); filled diamonds: \texttt{gpt-oss-20b} ($N{=}8$); hollow triangles/diamonds: Nemotron-Super/\texttt{gpt-oss-20b} ($N{=}3$, forced $\hat{\rho}{=}-0.500$). All 19 computable points show $\hat{\rho}<0$. Mean $\hat{\rho}=-0.122$ for $N{\geq}7$. No correlation headroom for diversity strategies.}
\label{fig:rho}
\end{figure}

This gives 19 computable points (Figure~\ref{fig:rho}). All nineteen are negative. The seven estimates from $N\geq7$ are $\hat{\rho}\in\{-0.167,-0.167,-0.143,-0.143,-0.100,-0.067,-0.067\}$, mean $-0.122$; two of these come from \texttt{gpt-oss-20b} at $N{=}8$. One additional point at $N{=}5$ (\texttt{gpt-oss-20b}, $\hat{\rho}=-0.250$) is excluded from the large-$N$ mean but confirms the sign. The eleven $N{=}3$ points (3 Nemotron + 8 \texttt{gpt-oss-20b}) all give $\hat{\rho}=-0.500$, which is the mathematical minimum for $N{=}3$ with $v_c\in\{1,2\}$; they confirm sign but not magnitude. Across all 19 points the mean is $-0.348$. When the model is wrong, wrong answers scatter across many distinct values rather than clustering, so two draws are less likely to share the correct answer than $\hat{p}^2$ predicts. This leaves no correlation headroom for diversity strategies to exploit. Diverse prompts cannot decorrelate what is already at or below zero.

\subsection{Weaker Strategies Reduce Accuracy}

The Mixer modifies two parameters simultaneously: it \textit{reduces} $\rho$ (intended) but also \textit{reduces} $\bar{p}$ (unintended: weaker prompts lower per-attempt accuracy). For the intervention to succeed, decorrelation must exceed accuracy loss.

The critical test is E10: Equal mix (2+2+2+2) at $T{=}0.5$, where prompt diversity should help most (low stochastic diversity). It scores 36, \textit{worse} than $T{=}0.5$ alone (38). Prompt diversity does not compensate even when temperature-based diversity is deliberately suppressed.

\subsection{Model Capability Dominates}

The cleanest comparison is \texttt{gpt-oss-20b} at $N{=}8$ (Table~\ref{tab:models}): identical inference budget as \model{}, same pipeline, same early stopping. Local: 9/10 vs.\ 10/10. LB: 31.0 (runs: 35, 28, 30\footnote{\label{fn:public}Third run from a public notebook with identical configuration (\texttt{gpt-oss-20b}, $N{=}8$, $T{=}1.0$, early\_stop$=$4). See \url{https://www.kaggle.com/code/nguyennguyen599/aimo3-skills-optional-luck-required-gpt-oss-20b}.}) vs.\ 39.3. The 8-point gap at equal $N$ is $4\times$ larger than any prompt optimization ($\pm$2 points). Scaling $N$ further does not compensate: \texttt{gpt-oss-20b} at $N{=}32$ scores 26, \textit{worse} than $N{=}8$ (31.0), because per-attempt time shrinks and $\hat{p}$ drops from 0.61 to 0.52.

\nemotron{}'s 23/50 ($N{=}3$) is confounded by compute budget. On HMMT Feb25, \nemotron{} NVFP4 scores 95.4\% vs.\ 94.7\% BF16; quantization is not the cause. \qwen{}'s 23/50 reflects $0.6\times$ active parameters.

Independent $\text{pass@}k$ data from the competition hosts\footnote{\label{fn:aimo-x}\url{https://x.com/AIMOprize/status/2039441022996934783}} confirms this. \model{} outperforms Nemotron-Cascade-2-30B-A3B at every $k \in \{1, 3, 5, 20, 100\}$ on both public and private sets, despite the latter claiming 2025 IMO Gold Medal-level performance. Benchmark capability on one distribution does not transfer.

\begin{table}[htbp]
\centering\small
\begin{tabular}{lcccccc}
\toprule
Model & Total / Active & Quant & $N$ & $\hat{p}$ & Local /10 & LB Score \\
\midrule
\model{} & 116.8B / 5.1B & FP8 & 8 & 0.69 & 10 & 39.3 (21 runs) \\
\texttt{gpt-oss-20b} & 20B / 3.6B & FP8 & 8 & 0.61 & 9 & 31.0 (3-run: 35,28,30\textsuperscript{\ref{fn:public}}) \\
\texttt{gpt-oss-20b} & 20B / 3.6B & FP8 & 32 & 0.52 & --- & 26 \\
\texttt{gpt-oss-20b} & 20B / 3.6B & FP8 & 3 & 0.54 & --- & 28.3 (3-run: 26,28,31) \\
\nemotron{}-NVFP4 & 120B / 12B & NVFP4 & 3 & 0.47 & --- & 23 \\
\qwen{} & 35B / 3B & BF16 & 16 & 0.46 & 8 & 23 \\
Nemotron-14B & 14B / 14B & BF16 & 8 & 0.03\textsuperscript{*} & --- & 3 \\
\bottomrule
\end{tabular}
\vspace{1mm}
\small \textsuperscript{*}Format mismatch; not representative of model capability.
\caption{Model comparison. At equal $N{=}8$, the 8-point gap between \model{} (39.3) and \texttt{gpt-oss-20b} (31.0) dwarfs all prompt optimizations ($\pm$2 points). Scaling $N$ to 32 on \texttt{gpt-oss-20b} backfires: per-attempt time shrinks, $\hat{p}$ drops from 0.61 to 0.52, score drops from 31.0 to 26.}
\label{tab:models}
\end{table}

The main result holds at every $N$: no inference-time optimization improved over baseline \textit{within} a fixed model, and the model capability gap persists regardless of compute budget.

\FloatBarrier
\section{Cross-Model Validation}

Two additional models test whether the result is model-specific.

\begin{figure}[htbp]
\centering
\includegraphics[width=0.80\linewidth]{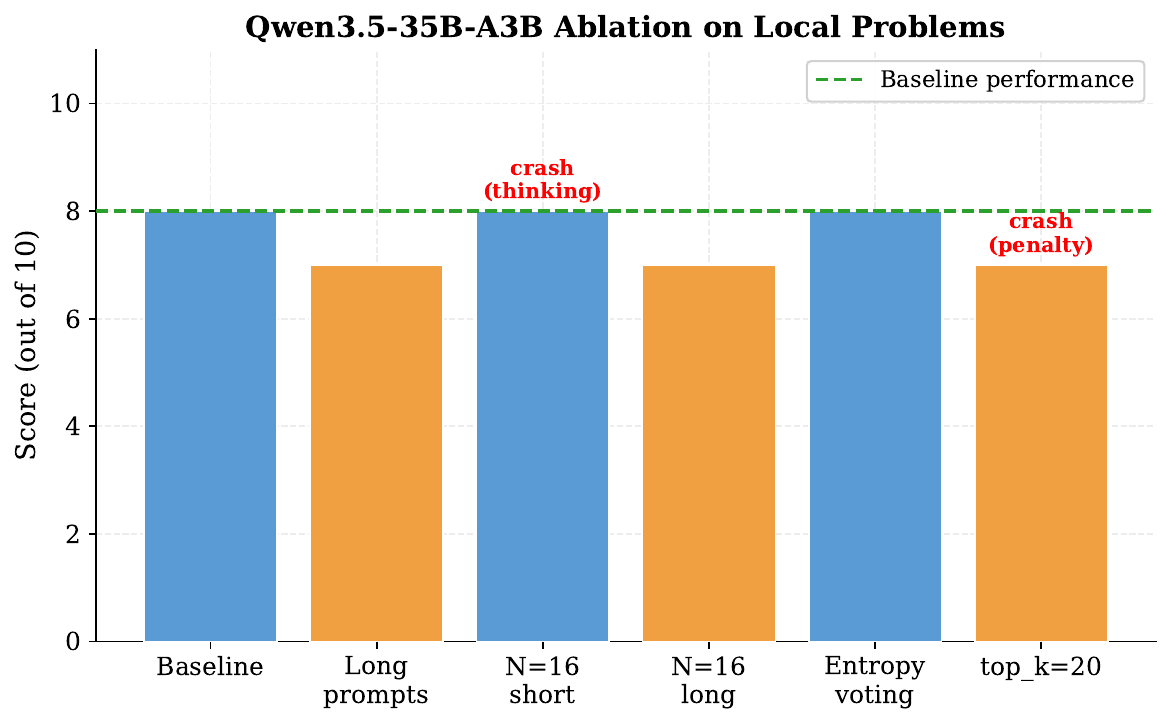}
\caption{\qwen{} ablation on 10 local problems. Blue: baseline ($8/10$). Orange: underperform ($7/10$). Red: crashed. Nothing improves beyond baseline.}
\label{fig:qwen}
\end{figure}

\paragraph{Qwen3.5-35B-A3B.} 35B total, 3B active via sparse MoE with Gated Delta Networks \citep{yang2025gateddeltatnet, qwen2026qwen35}. Eight controlled experiments on 10 local problems, each changing exactly one variable (Table~\ref{tab:qwen}, Figure~\ref{fig:qwen}). Doubling $N$ from 8 to 16: no improvement. Long prompts: $-1$ point. Manufacturer-recommended parameters: $-1$ or crash. LB submission: 23/50.

\begin{table}[htbp]
\centering
\begin{tabular}{llcc}
\toprule
Experiment & Change & $N$ & Local Score \\
\midrule
Baseline & --- & 8 & 8/10 \\
Long prompts & From GPT-OSS & 16 & 7/10 \\
$N{=}16$, short & Double attempts & 16 & 8/10 \\
$N{=}16$, long & Double + long prompts & 16 & 7/10 \\
Entropy voting & Add logprobs$=$5 & 16 & 8/10 \\
\texttt{top\_k=20} & Qwen recommended & 16 & 7/10 \\
Thinking mode & \texttt{enable\_thinking} & 8 & \textcolor{red}{crash} \\
\texttt{presence\_penalty} & Qwen recommended & 16 & \textcolor{red}{crash} \\
\bottomrule
\end{tabular}
\caption{Qwen3.5-35B-A3B ablation. Nothing improves beyond baseline.}
\label{tab:qwen}
\end{table}

\paragraph{Nemotron-Super-120B-NVFP4.} 120B total, 12B active via hybrid Mamba-2 + MoE + Attention \citep{moshkov2025aimo2}, NVFP4 quantized. $N{=}3$ attempts with tool-integrated reasoning, fp8 KV cache, 49K-token context. Local test: 6--9/10 across configurations. LB: 23/50, identical to Qwen despite $4\times$ active parameters. On HMMT Feb25, NVFP4 scores 95.4\% vs.\ 94.7\% BF16; quantization is not the cause. The comparison is confounded by $N{=}3$ vs.\ \model{}'s $N{=}8$.

\FloatBarrier
\section{Complete Ablation}

Twenty-three experiments on \model{}, each modifying one variable. None reliably improve over baseline.

\begin{table}[H]
\centering
\small
\begin{tabular}{llccc}
\toprule
ID & Change & LB Score & $\Delta$ & Category \\
\midrule
Baseline & --- (21-run mean) & 39.3 & --- & --- \\
\midrule
\#1 & Conservative mix (5+1+1+1) & 40 & +0.7 & Diversity \\
\#2 & Aggressive mix (3+2+2+1) & 40 & +0.7 & Diversity \\
\#3 & Equal mix (2+2+2+2) & 38 & $-$1.3 & Diversity \\
E1 & 8$\times$ Small Cases & 37 & $-$2.3 & Strategy \\
E2 & 8$\times$ Work Backwards & 39 & $-$0.3 & Strategy \\
E3 & 8$\times$ Classify & 36 & $-$3.3 & Strategy \\
E4 & $T{=}0.5$ & 38 & $-$1.3 & Temperature \\
E5 & $T{=}0.8$ & 40 & +0.7 & Temperature \\
E6 & $T{=}1.2$, min\_p$=$0.03 & 37 & $-$2.3 & Temperature \\
E10 & Equal mix @ $T{=}0.5$ & 36 & $-$3.3 & Diversity+T \\
E12 & 8$\times$ Code-first & 41 & +1.7 & Strategy \\
C.1 & Code-first confirm & 38 & $-$1.3 & Confirm \\
C.2 & Code-first confirm & 34 & $-$5.3 & Confirm \\
EF1 & Formalize-First (F-1) & 39 & $-$0.3 & Strategy \\
E11 & $N{=}3$ (elaborate prompt) & 36 & $-$3.3 & N-ablation \\
\midrule
\multicolumn{5}{l}{\textit{Cross-model (gpt-oss-20b):}} \\
G.N8 & gpt-oss-20b, $N{=}8$ & 31.0 & $-$8.3 & Model \\
G.N32 & gpt-oss-20b, $N{=}32$ & 26 & $-$13.3 & Model \\
\multicolumn{5}{l}{\textit{Cross-model (other):}} \\
EN1 & Nemotron-Super-120B, $N{=}3$ & 23 & $-$16.3 & Model \\
\bottomrule
\end{tabular}
\caption{Complete ablation on \model{}. No experiment reliably exceeds baseline mean. E12 (41) was not replicated in confirmation runs. \texttt{gpt-oss-20b} at equal $N{=}8$ scores 31.0; scaling to $N{=}32$ drops to 26 as per-attempt time shrinks.}
\label{tab:ablation}
\end{table}

\begin{figure}[H]
\centering
\includegraphics[width=0.95\linewidth]{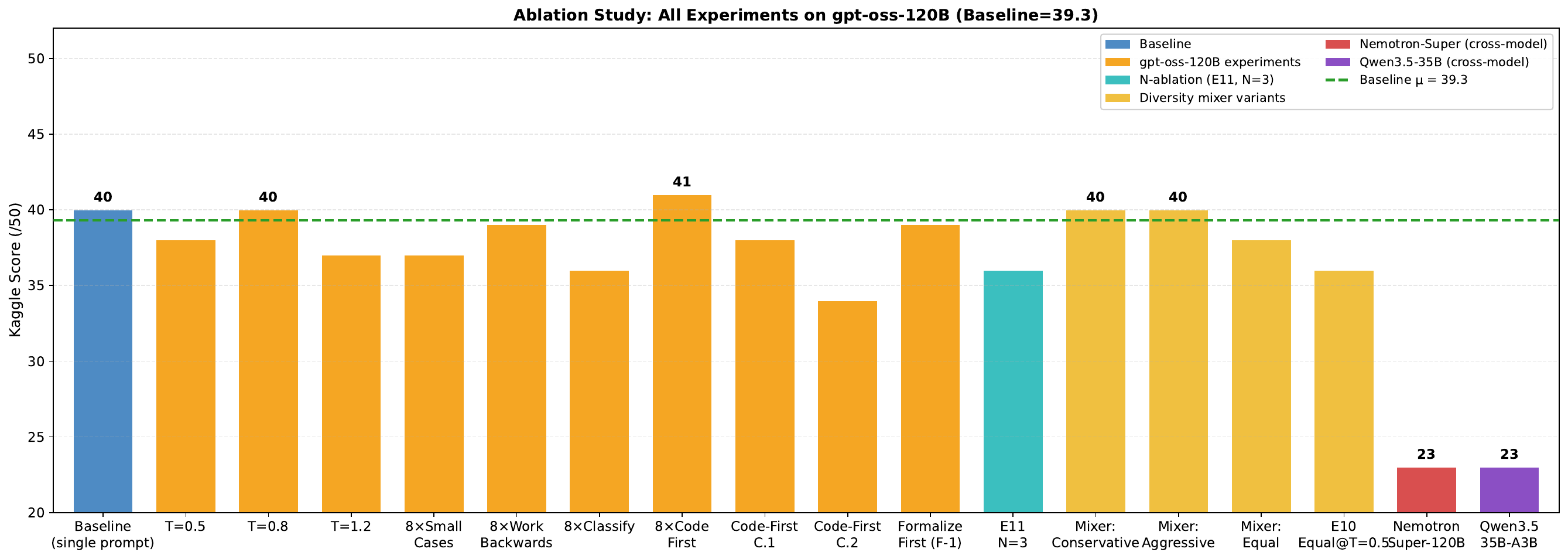}
\caption{Complete ablation across all experiments. Blue: baseline (39.3). Orange: interventions. Yellow-orange: diversity mixer. Teal: N-ablation. Red/Purple: cross-model. Green dashed: baseline mean. No experiment reliably exceeds baseline.}
\label{fig:waterfall}
\end{figure}

The optimization landscape is flat (Figure~\ref{fig:waterfall}). The best experiment (E12 Code-first, 41/50) fell to 38 and then 34 on two confirmation runs, yielding a 3-run mean of 37.7, below baseline mean of 39.3. The Formalize-First prompt (F-1), which forces explicit equation formulation before computation, scored 39. The original system is a local optimum.

\FloatBarrier
\section{Comparison with State of the Art}
\label{sec:sota}

\subsection{AIMO Competition Progression}

\begin{table}[H]
\centering
\begin{tabular}{lcccp{4.8cm}}
\toprule
Competition & Winner & This Work & $N$ & Key Innovation \\
\midrule
AIMO-1 (2024) & 29/50 & --- & 48 & Brute-force voting \citep{ai-mathematical-olympiad-prize,aimo1winner} \\
AIMO-2 (2025) & 34/50 & --- & 64 & Custom training + GenSelect \citep{ai-mathematical-olympiad-progress-prize-2,moshkov2025aimo2} \\
AIMO-3 (2026) & 46\textsuperscript{*} & 42 & 8 & Off-the-shelf model \citep{ai-mathematical-olympiad-progress-prize-3} \\
\bottomrule
\end{tabular}
\caption{Competition progression. From high-$N$ to high-$\hat{p}$: as models improve, inference-time tricks yield diminishing returns. \textsuperscript{*}Top leaderboard score.}
\label{tab:progression}
\end{table}

\begin{figure}[H]
\centering
\includegraphics[width=0.82\linewidth]{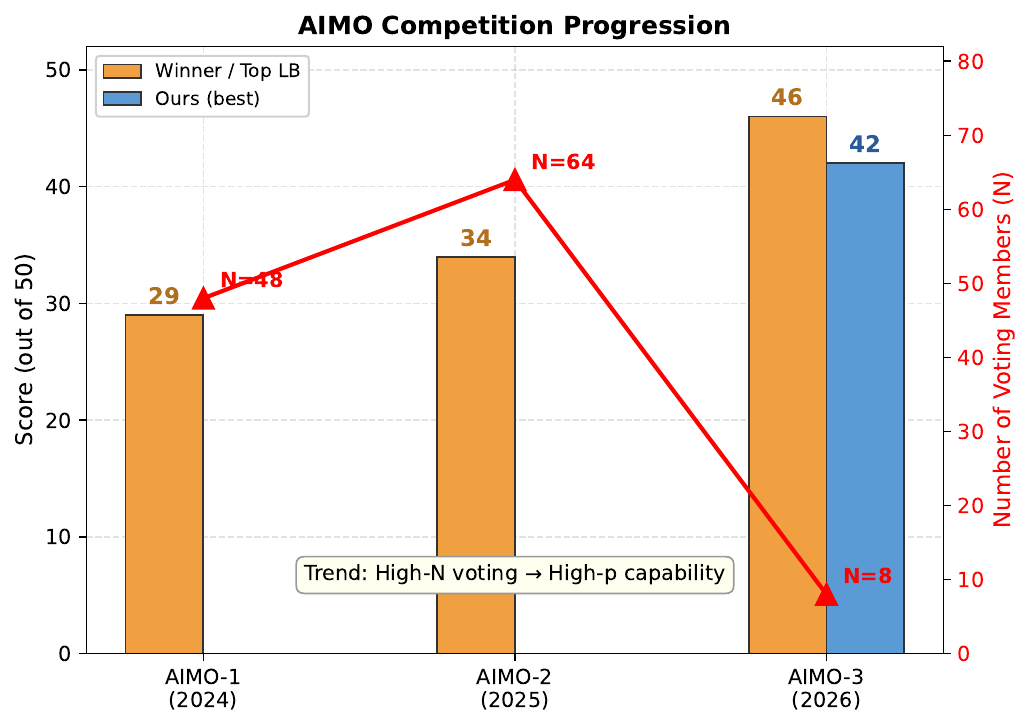}
\caption{AIMO competition progression. Orange bars: winner/top LB scores (29$\to$34$\to$46). Blue bar: this work (42). Red line: voters $N$ (48$\to$64$\to$8). High-$N$ voting gives way to high-$\hat{p}$ capability.}
\label{fig:progression}
\end{figure}

The trend: from high-$N$ to high-$p$ (Figure~\ref{fig:progression}). AIMO-1 used $N{=}48$ with brute-force voting. AIMO-2 invested in 540K training problems with custom GenSelect. This system uses zero training compute on an off-the-shelf model. Model capability first, everything else second.

\subsection{Inference-Time Scaling}

Scaling test-time compute \citep{snell2024scalingllmtesttime, brown2024largelanguagemonkeys} can substitute for model scaling. The present results qualify this: on IMO-level problems at $\hat{p} \approx 0.69$, repeated sampling works, but \textit{modifying} the sampling (prompt diversity, temperature tuning, strategy mixing) does not improve over vanilla self-consistency. Returns to inference-time optimization flatten once the base system is configured.

The OpenAI $\times$ AIMO evaluation (March 2025) showed commercial models solving 50/50 AIMO-2 problems with sufficient compute; the highest open-source Kaggle score was 34/50 \citep{aimoprize2025gap}. The gpt-oss-120b baseline (39.3 mean, 42 best) narrows this gap with zero training compute. The top AIMO-3 score (46+) indicates further room.

\FloatBarrier
\section{Submission as Lottery}

\begin{figure}[H]
\centering
\includegraphics[width=0.95\linewidth]{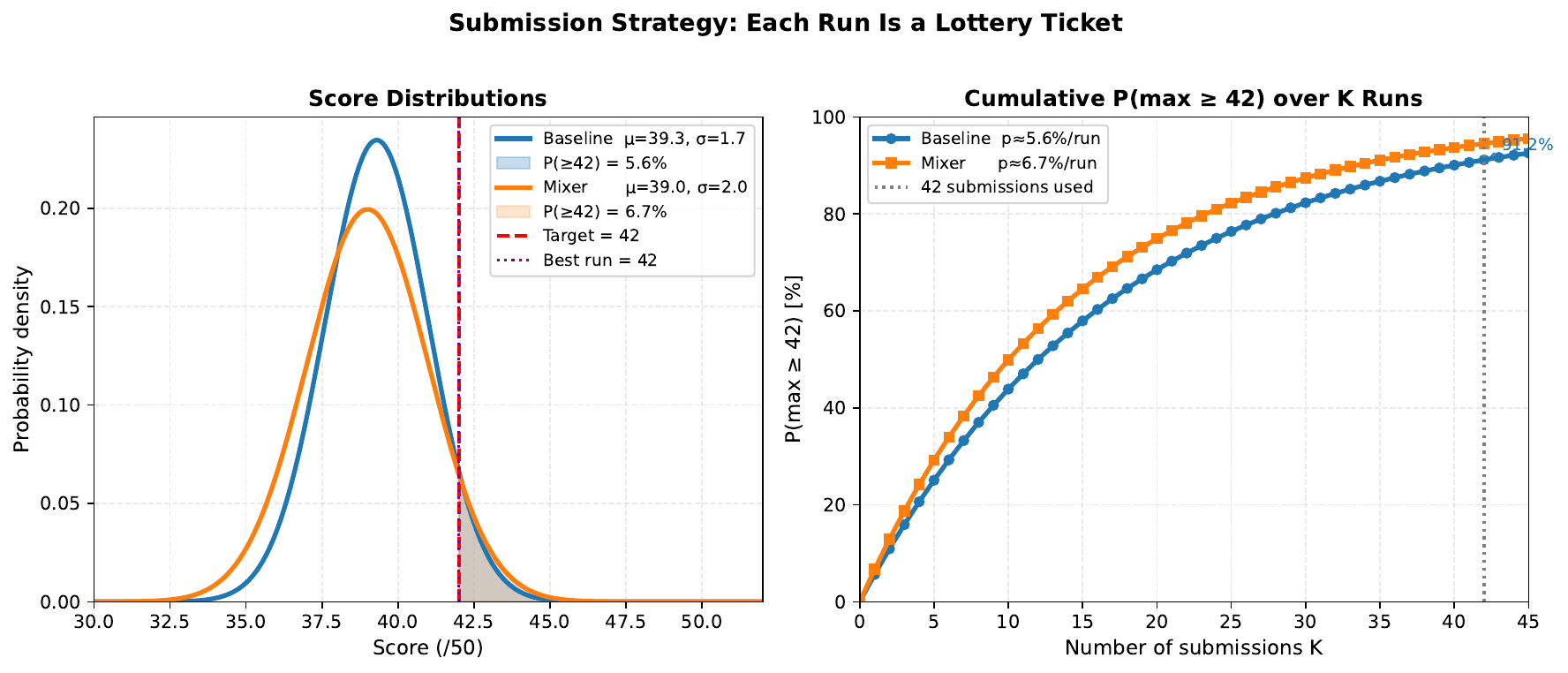}
\caption{Left: score distributions. Baseline ($\mu{=}39.3$, $\sigma{=}1.7$, blue) vs.\ Mixer ($\mu{\approx}39.0$, $\sigma{\approx}2.0$, orange). Red dashed line: target score 42. Right: cumulative probability of $\max \geq 42$ over $K$ submissions. Baseline: $p{\approx}0.056$ per run. Mixer: $p{\approx}0.037$. Dotted line at 42 submissions used.}
\label{fig:lottery}
\end{figure}

With $\hat{\mu} = 39.3$ and $\hat{\sigma} = 1.7$ over 21 baseline runs (range 34--42), each submission is a lottery ticket. The best run reached 42/50; each has $P(\text{score} \geq 42) \approx 5.6\%$. The Mixer reduces expected score ($\mu$ drops to ${\sim}39.0$) without improving tail probability. The optimal strategy: submit the unmodified baseline repeatedly (Figure~\ref{fig:lottery}). All 42 submissions exhausted.

Part of the variance is infrastructure noise: shared-GPU benchmarks can shift by ${\sim}6$ percentage points from resource contention alone \citep{anthropic2026infranoise}. The 21-run baseline and 3-model cross-validation mitigate this.

\section{Selection Loss}
\label{sec:selection}

A host-posted analysis\footnote{\url{https://www.kaggle.com/competitions/ai-mathematical-olympiad-progress-prize-3/discussion/679559}}\textsuperscript{,\ref{fn:aimo-x}} reports \model{} at $\text{pass@}20 \approx 45.5$ on the AIMO~3 private set (95\% bootstrap CI $[43, 48]$) and $\text{pass@}100 \approx 49/50$. Only one problem across both public and private sets remains unsolved out of the box. The raw ceiling sits above every score in the ablation; the negative results therefore bound prompt-level interventions inside a fixed majority-voting selector, not the model's capability.

Six points separate the best majority-vote score (42) from the $\text{pass@}20$ mean (45.5). Equation~\ref{eq:majority} already absorbs $p$ and $\rho$, so the gap is neither. It is selection loss: the correct answer appears in the $N{=}8$ pool but is outvoted by a more common wrong one. Majority voting is the cheapest possible selector. A verifier that distinguishes right from wrong without ground truth (code execution against problem constraints, formal substitution, cross-candidate consistency) would capture some of those six points. This paper does not test one.

The scope of the negative result is narrower than the title suggests. Prompt-level inference-time optimization does not help. Selection-level optimization is a separate question and remains open.

\section{Conclusion}

Diverse Prompt Mixer was designed to decorrelate errors in majority voting for mathematical reasoning. It does not work. High-temperature sampling already provides sufficient diversity; structured prompt diversity is redundant at best, harmful at worst. Across 3 models, 23+ experiments, and 50 IMO-level problems, model capability dominates all \textit{prompt-level} inference-time optimizations by $4\times$ (8-point gap vs.\ $\pm$2-point prompt effects at equal $N{=}8$). Scaling $N$ past the compute budget is counterproductive. For hardware-constrained competitions: use the largest model that fits, keep temperature high, and spend submission budget on lottery tickets, not prompt engineering. Six points separate the best majority-vote run from $\text{pass@}20$ (Section~\ref{sec:selection}). A verifier-based selector is the direction left open.

\section*{Acknowledgments}
Thanks to the AIMO Prize Committee and Kaggle for organizing the competition and providing GPU infrastructure. The baseline notebook builds on the inference pipeline originally developed by nihilisticneuralnet.\footnote{\url{https://www.kaggle.com/code/nihilisticneuralnet/44-50-let-me-over-cook}}

\bibliographystyle{plainnat}
\bibliography{references}

\appendix
\setcounter{table}{0}
\renewcommand{\thetable}{A\arabic{table}}
\setcounter{figure}{0}
\renewcommand{\thefigure}{A\arabic{figure}}

\section{Detailed System Configuration}
\label{app:config}

\begin{table}[H]
\centering
\small
\begin{tabular}{lll}
\toprule
Parameter & Value & Notes \\
\midrule
\multicolumn{3}{l}{\textbf{Model \& Serving}} \\
Model & \model{} & 116.8B total, 5.1B active (MoE) \\
Model path & \texttt{/kaggle/input/gpt-oss-120b/} & Kaggle model hub \\
Quantization & FP8 (weights + KV cache) & \texttt{kv\_cache\_dtype=fp8\_e4m3} \\
Engine & vLLM 0.11.x & \texttt{pip install vllm==0.11.2} \\
GPU & NVIDIA \gpu{} & Kaggle competition kernel \\
Tensor parallel & 1 & Single GPU \\
GPU memory utilization & 0.96 & 76.8\,GB of 80\,GB \\
Max batch size & 256 & vLLM \texttt{--max-num-seqs} \\
\midrule
\multicolumn{3}{l}{\textbf{Sampling}} \\
Temperature & 1.0 & Optimal (Section 4.1) \\
min\_p & 0.02 & Nucleus-like filtering \\
Context tokens & 65536 & Max sequence length \\
Buffer tokens & 512 & Reserved for output \\
Top logprobs & 5 & For entropy computation \\
Attempts ($N$) & 8 & Parallel per problem \\
Max turns & 128 & Tool-call rounds per attempt \\
Early stop & 4/8 agreement & Non-trivial answers only \\
Seed & 42 & Base seed; per-attempt varies \\
\midrule
\multicolumn{3}{l}{\textbf{Voting}} \\
Method & Entropy-weighted majority & \\
Weight & $w = 1 + 1/(\text{entropy} + 0.1)$ & Confident $\rightarrow$ higher weight \\
Tie-breaking & Highest total weight & \\
\midrule
\multicolumn{3}{l}{\textbf{Sandbox}} \\
Kernels & 8 persistent Jupyter kernels & One per attempt \\
Jupyter timeout & 6\,s per execution & Prevents infinite loops \\
Sandbox timeout & 3\,s to acquire kernel & \\
Libraries & sympy, numpy, mpmath, itertools & Pre-installed \\
\midrule
\multicolumn{3}{l}{\textbf{Time Budget}} \\
Total Kaggle limit & 18000\,s (5 hours) & Hard wall-clock limit \\
Infrastructure buffer & 540\,s & 9\,min for variance \\
Startup budget & 360\,s & Preload + vLLM + kernels \\
Solving budget & 17100\,s & For 50 problems \\
Base problem timeout & 300\,s & Default per-problem \\
High problem timeout & 900\,s & Hard cap per problem \\
Session timeout & 960\,s & OpenAI client timeout \\
\bottomrule
\end{tabular}
\caption{Complete baseline configuration.}
\label{tab:config}
\end{table}

\section{System Prompts}
\label{app:prompts}

Full text of all system prompts used in experiments. Each experiment changes only the system prompt; all other parameters remain as in Table~\ref{tab:config}.

\subsection{Original (Baseline)}
\begin{lstlisting}[language={},title={system\_prompt}]
You are an elite mathematical problem solver with expertise
at the International Mathematical Olympiad (IMO) level.
Your goal is to find the correct answer through rigorous
mathematical reasoning.

# Problem-Solving Approach:
1. UNDERSTAND: Carefully read and rephrase the problem in
   your own words. Identify what is given, what needs to
   be found, and any constraints.
2. EXPLORE: Consider multiple solution strategies. Think
   about relevant theorems, techniques, patterns, or
   analogous problems. Don't commit to one approach
   immediately.
3. PLAN: Select the most promising approach and outline
   key steps before executing.
4. EXECUTE: Work through your solution methodically.
   Show all reasoning steps clearly.
5. VERIFY: Check your answer by substituting back,
   testing edge cases, or using alternative methods.
   Ensure logical consistency throughout.

# Mathematical Reasoning Principles:
- Break complex problems into smaller, manageable
  sub-problems
- Look for patterns, symmetries, and special cases
  that provide insight
- Use concrete examples to build intuition before
  generalizing
- Consider extreme cases and boundary conditions
- If stuck, try working backwards from the desired result
- Be willing to restart with a different approach if
  needed

# Verification Requirements:
- Cross-check arithmetic and algebraic manipulations
- Verify that your solution satisfies all problem
  constraints
- Test your answer with simple cases or special values
  when possible
- Ensure dimensional consistency and reasonableness
  of the result

# Output Format:
The final answer must be a non-negative integer between
0 and 99999.
Place your final numerical answer inside \boxed{},
e.g., \boxed{42}

Think step-by-step and show your complete reasoning
process. Quality of reasoning is as important as the
final answer.
\end{lstlisting}

\subsection{Small Cases First (E1)}
\begin{lstlisting}[language={},title={system\_prompt}]
You are an elite IMO-level problem solver. Your primary
strategy is to start with small cases.

1. ENUMERATE: Compute the answer for n=1,2,3,4,5,...
   using code or by hand.
2. PATTERN: Look for a pattern in the small cases.
   Can you find a recurrence? A closed form?
3. CONJECTURE: State your conjecture precisely.
4. PROVE: Prove the conjecture holds in general, or
   compute the answer directly from the pattern.
5. VERIFY: Check with an independent method.

Place your final answer inside \boxed{}.
\end{lstlisting}

\subsection{Work Backwards (E2)}
\begin{lstlisting}[language={},title={system\_prompt}]
You are an elite IMO-level problem solver. Your primary
strategy is to work backwards from the answer.

1. CONSTRAINTS: List all constraints the answer must
   satisfy.
2. NARROW: What properties must the solution have?
   Eliminate impossibilities.
3. CONSTRUCT: Build the answer from the constraints.
4. VERIFY: Check all constraints are satisfied.

Place your final answer inside \boxed{}.
\end{lstlisting}

\subsection{Classify Then Solve (E3)}
\begin{lstlisting}[language={},title={system\_prompt}]
You are an elite IMO-level problem solver. First
classify, then solve.

1. CLASSIFY: Is this number theory, algebra,
   combinatorics, or geometry?
2. RECALL: What canonical techniques apply to this
   type?
3. APPLY: Use the most relevant technique.
4. VERIFY: Check your answer.

Place your final answer inside \boxed{}.
\end{lstlisting}

\subsection{Code-First (E12)}
\begin{lstlisting}[language={},title={system\_prompt}]
You are an elite IMO-level problem solver. Always
start with code.

1. IMPLEMENT: Write Python code to explore the problem
   computationally. Start with brute-force for small
   cases.
2. OBSERVE: What do the computational results tell you?
3. GENERALIZE: Find the pattern or formula.
4. COMPUTE: Calculate the final answer.
5. VERIFY: Cross-check with an independent method.

Place your final answer inside \boxed{}.
\end{lstlisting}

\subsection{Formalize-First / F-1 (EF1)}
\begin{lstlisting}[language={},title={system\_prompt}]
Before writing any code, formalize the problem:

1. Define variables: "Let n = ..., let f(x) = ..."
2. State constraints as equations
3. Identify the objective: "Find: max(f(n)) mod 1000"

Then implement code that solves your equations.
Verify your answer. Place it inside \boxed{}.
\end{lstlisting}

\subsection{Preference Prompt (appended to every problem)}
\begin{lstlisting}[language={},title={preference\_prompt (appended to user message)}]
You have access to `math`, `numpy`, and `sympy` for:

# Symbolic Computation (sympy):
- Algebraic manipulation and simplification
- Solving equations and systems of equations
- Number theory functions (primes, divisors, modular
  arithmetic)
- Polynomial operations and factorization

# Numerical Computation (numpy):
- Array operations and linear algebra
- Efficient numerical calculations

Best Practices:
- Use sympy for exact symbolic answers when possible
- Use numpy for numerical verification
- Combine symbolic and numerical approaches
- Validate results against known cases
\end{lstlisting}

\section{vLLM Server Configuration}
\label{app:vllm}

Exact command used to launch the inference server:

\begin{lstlisting}[language=bash,title={vLLM launch command}]
python -m vllm.entrypoints.openai.api_server \
  --seed 42 \
  --model /kaggle/input/gpt-oss-120b/transformers/default/1 \
  --served-model-name gpt-oss \
  --tensor-parallel-size 1 \
  --max-num-seqs 256 \
  --gpu-memory-utilization 0.96 \
  --kv-cache-dtype fp8_e4m3 \
  --dtype auto \
  --max-model-len 65536 \
  --enable-auto-tool-choice \
  --tool-call-parser pythonic \
  --port 8000
\end{lstlisting}

\paragraph{Dependencies (exact versions):}\leavevmode
\begin{lstlisting}[language=bash]
pip install vllm==0.11.2 openai sympy numpy mpmath \
    polars kaggle-evaluation jupyter_client
\end{lstlisting}

\section{Complete Submission Log}
\label{app:submissions}

All 42 submissions. One daily submission allowed.

\begin{table}[H]
\centering
\footnotesize
\begin{tabular}{clcl}
\toprule
\# & Configuration & LB & Notes \\
\midrule
0 & Baseline 8$\times$ original & 42 & Best ever \\
1 & Conservative (5+1+1+1) & 40 & \\
2 & Aggressive (3+2+2+1) & 40 & \\
3 & Equal (2+2+2+2) & 38 & Worst diversity \\
4 & Nemotron-14B ($N$=16) & 3 & Format mismatch \\
5 & Qwen3.5-35B-A3B & 23 & Cross-model \\
6 & Baseline validation & 39 & \\
7 & Baseline validation & 37 & \\
8 & Baseline validation & 40 & \\
9 & Baseline validation & 40 & \\
10 & Baseline validation & 39 & \\
11 & E1: 8$\times$ Small Cases & 37 & \\
12 & E2: 8$\times$ Backwards & 39 & \\
13 & E3: 8$\times$ Classify & 36 & Weakest \\
14 & E4: $T$=0.5 & 38 & \\
15 & E5: $T$=0.8 & 40 & \\
16 & E6: $T$=1.2 & 37 & \\
17 & E10: Equal mix @ $T$=0.5 & 36 & Critical test \\
18 & E12: 8$\times$ Code-first & 41 & \\
20 & Code-first confirm (C.1) & 38 & \\
21 & Code-first confirm (C.2) & 34 & 3-run mean = 37.7 \\
22 & Lucky run \#1 & 39 & \\
23 & Lucky run \#2 & 39 & \\
24 & Lucky run \#3 & 41 & \\
25 & Lucky run \#4 & 40 & \\
EN1 & Nemotron-Super LB & 23 & Cross-model \\
EF1 & Formalize-First (F-1) & 39 & \\
E11 & $N{=}3$ + elaborate prompt & 36 & gpt-oss N=3=36 vs Nemotron N=3=23 \\
\midrule
26 & Lucky run \#5 & 40 & \\
27 & Lucky run \#6 & 38 & \\
28 & Lucky run \#7 & 39 & \\
29 & Lucky run \#8 & 39 & \\
30 & Lucky run \#9 & 39 & \\
31 & Lucky run \#10 & 34 & \\
32 & Lucky run \#11 & 40 & \\
33 & Lucky run \#12 & 42 & \\
\midrule
G1 & \texttt{gpt-oss-20b} $N{=}3$ run 2 & 31 & Cross-model \\
G2 & \texttt{gpt-oss-20b} $N{=}3$ run 3 & 28 & N=3 mean: 28.3 \\
G3 & \texttt{gpt-oss-20b} $N{=}32$ & 26 & Worse than N=8 \\
G4 & \texttt{gpt-oss-20b} $N{=}8$ run 1 & 35 & Equal-N comparison \\
G5 & \texttt{gpt-oss-20b} $N{=}8$ run 2 & 28 & \\
G6 & \texttt{gpt-oss-20b} $N{=}8$ (public\textsuperscript{$\dagger$}) & 30 & 3-run mean: 31.0 \\
\bottomrule
\end{tabular}
\caption{Complete submission log (42 entries). \textsuperscript{$\dagger$}Public notebook with identical config.}
\label{tab:submissions}
\end{table}

\section{Baseline Score Distribution}
\label{app:baseline}

Twenty-one baseline runs (8$\times$ original, $T{=}1.0$, all identical configuration):

\begin{center}
\{42, 42, 41, 40, 40, 40, 40, 40, 40, 40, 39, 39, 39, 39, 39, 39, 39, 38, 38, 37, 34\}
\end{center}

\noindent $\hat{\mu} = 39.3$, $\hat{\sigma} = 1.7$, min $= 34$, max $= 42$.

Per-attempt accuracy from $\mathbb{E}[\text{score}] = 50 \cdot P(X \geq 5)$, $X \sim \text{Binomial}(8, p)$: $\hat{p} \approx 0.69$.

\noindent Key probabilities (using Normal approximation $\mathcal{N}(39.3, 1.7^2)$):
\begin{itemize}[nosep]
    \item $P(\text{score} \geq 42 \mid \text{single run}) \approx 5.6\%$
    \item 42 submissions used
\end{itemize}

\section{Reproduction Guide}
\label{app:reproduce}

Step-by-step instructions to reproduce all results from scratch:

\begin{enumerate}
    \item \textbf{Environment}: Create a Kaggle notebook with GPU T4$\times$2 or P100 for local testing, or submit to AIMO~3 competition for \gpu{} evaluation.

    \item \textbf{Model}: Add \model{} from Kaggle Models\\ {\small\path{/kaggle/input/gpt-oss-120b/transformers/default/1}}.

    \item \textbf{Install dependencies}:
    \begin{lstlisting}[language=bash]
pip install vllm==0.11.2 openai sympy numpy mpmath \
    polars kaggle-evaluation jupyter_client
    \end{lstlisting}

    \item \textbf{Preload model weights} (reduces cold-start):
    \begin{lstlisting}[language=python]
# Read all model files into OS page cache
for root, _, files in os.walk(model_path):
    for f in files:
        with open(os.path.join(root, f), 'rb') as fh:
            while fh.read(1024*1024*1024): pass
    \end{lstlisting}

    \item \textbf{Start vLLM server}: Use exact flags from Appendix~\ref{app:vllm}.

    \item \textbf{Initialize Jupyter kernels}: Create 8 persistent kernels for parallel code execution.

    \item \textbf{Run inference}: The \texttt{predict()} function receives problems one at a time from Kaggle's evaluation server. Each call invokes \texttt{solver.solve\_problem()}.

    \item \textbf{Expected results}: Score 34--42 (mean 39.3) per run. Total runtime ${\sim}4.5$ hours.

    \item \textbf{Experiment variants}: To replicate any experiment, change only the system prompt (Appendix~\ref{app:prompts}) or the specific parameter noted in Table~\ref{tab:ablation}. All other configuration remains identical.
\end{enumerate}

\paragraph{Code availability.} All experiment notebooks are included in the supplementary materials attached to this writeup. Each notebook is self-contained and runnable on Kaggle.

\section{Cross-Model Configurations}
\label{app:crossmodel}

\begin{table}[H]
\centering
\small
\begin{tabular}{lll}
\toprule
Parameter & Qwen3.5-35B-A3B & Nemotron-Super-120B \\
\midrule
Total params & 35B & 120B \\
Active params & 3B (MoE) & 12B (MoE+Mamba-2) \\
Quantization & BF16 & NVFP4 \\
Engine & SGLang & vLLM 0.17.0rc0 \\
$N$ (attempts) & 8 / 16 & 3 / 4 \\
Temperature & 1.0 & 1.0 \\
Context tokens & 32768 & 49152 \\
KV cache & Standard & FP8 \\
GPU memory & ${\sim}70$\,GB & ${\sim}67$\,GB \\
LB Score & 23/50 & 23/50 \\
\bottomrule
\end{tabular}
\caption{Cross-model validation configurations.}
\label{tab:crossmodel}
\end{table}

\end{document}